%% file: main.tex
\documentclass{Interspeech}

\interspeechcameraready  %

\usepackage{microtype}
\usepackage{xcolor}
\usepackage{xspace}

\usepackage{pifont} %

\usepackage{booktabs}
\usepackage{multirow}
\usepackage{tabularx}

\newcolumntype{C}{>{\centering\arraybackslash}X}
\newcolumntype{L}{>{\raggedright\arraybackslash}X}
\newcolumntype{R}{>{\raggedleft\arraybackslash}X}
\newcolumntype{P}[1]{>{\raggedright\arraybackslash}p{#1}}
\usepackage{etoolbox}                           %
\newcommand{\ubold}{\fontseries{b}\selectfont}  %
\robustify\ubold                                %

\graphicspath{{imgs/}}
\usepackage{tikz}
\usetikzlibrary{backgrounds}
\usetikzlibrary{fit}
\usetikzlibrary{calc}
\usetikzlibrary{shapes.geometric}

\usepackage{amsmath}

\usepackage{cite}
\newcommand{\etal}{\textit{et al.}\xspace}

\newcommand{\word}[1]{\textit{#1}}

\newcommand{\ab}{\mathbf{a}}

\newcommand{\ib}{\mathbf{i}}

\usepackage{manfnt}

\title{The mutual exclusivity bias of bilingual visually grounded speech models}
\author[affiliation={1}]{Dan}{Oneață}
\author[affiliation={2}]{Leanne}{Nortje}
\author[affiliation={3}]{Yevgen}{Matusevych}
\author[affiliation={2}]{Herman}{Kamper}

\affiliation{SpeeD Lab}{Politehnica Bucharest}{Romania}
\affiliation{Electrical and Electronic Engineering}{Stellenbosch University}{South Africa}
\affiliation{CLCG}{University of Groningen}{the Netherlands}

\email{\footnotesize \{dan.oneata,nortjeleanne,kamperh\}@gmail.com \ \ yevgen.matusevych@rug.nl}

\usepackage[prependcaption,textsize=scriptsize]{todonotes}
\definecolor{mycolor}{HTML}{008000}%
\definecolor{indiagreen}{HTML}{138808}%
\definecolor{papaya}{HTML}{EE892F}%
\definecolor{mygreen}{HTML}{008000}%
\definecolor{mypurple}{HTML}{9966CC}%
\definecolor{myblue}{HTML}{5D8AA8}
\definecolor{mypink}{HTML}{EC008C}

\newcommand{\new}[1]{#1}

\begin{document}

\maketitle

\begin{abstract}
Mutual exclusivity (ME) is a strategy where a novel word is associated with a novel object rather than a familiar one, facilitating language learning in children. Recent work has found an ME bias in a visually grounded speech (VGS) model trained on English speech with paired images. But ME has also been studied in bilingual children, who may employ it less due to cross-lingual ambiguity. We explore this pattern computationally using bilingual VGS models trained on combinations of English, French, and Dutch. We find that bilingual models generally exhibit a weaker ME bias than monolingual models, though exceptions exist. Analyses show that the combined visual embeddings of bilingual models have a smaller variance for familiar data, partly explaining the increase in confusion between novel and familiar concepts. We also provide new insights into why the ME bias exists in VGS models in the first place. %
\new{Code and data: \url{https://github.com/danoneata/me-vgs}.}
\end{abstract}

\section{Introduction}

The mutual exclusivity (ME) bias is a constraint that young children use in language learning, where they prefer to associate novel words with unfamiliar referents rather than familiar ones. 
For instance, if a child hears a novel word \word{aardvark} %
during book reading, they will naturally
map it to the unusual animal in the picture rather than the ordinary cat %
beside it.
This %
strategy enables efficient learning by narrowing down the space of possible referents \cite{markman_childrens_1988} and has been well-documented in children \cite{merriman_mutual_1989, markman_use_2003, lewis_role_2020}.

At the same time, the ME strategy might %
not apply to the same extent in bilingual situations, where each object can have more than one name. %
Bilingual children
therefore generally show a weaker ME bias compared to monolingual children \cite{byers2009, byers2017, houston2010}.
But this is not always the case, with results affected by
a child's age \cite{davidson2005, kalashnikova_effects_2015}, their vocabulary~\cite{byers2013}, the amount of time since their exposure to the familiar objects~\cite{kalashnikova2018}, and the type of test used~\cite{davidson1997, kalashnikova_effects_2015}. In short, the big picture of the differences in ME bias between monolingual and bilingual children is still not  clear.

\begin{figure}
    \centering
    \input{fig-overview}
    \caption{%
        Looking for ME in bilingual VGS models.
        We first train a model on images and spoken words from two languages (e.g., English and Dutch).
        We then test for ME by 
        pairing a novel word with a novel image and a familiar image.
        If the model gives a higher score to the novel--novel pair, then it has an ME bias.
    }
    \label{fig:overview}
\end{figure}
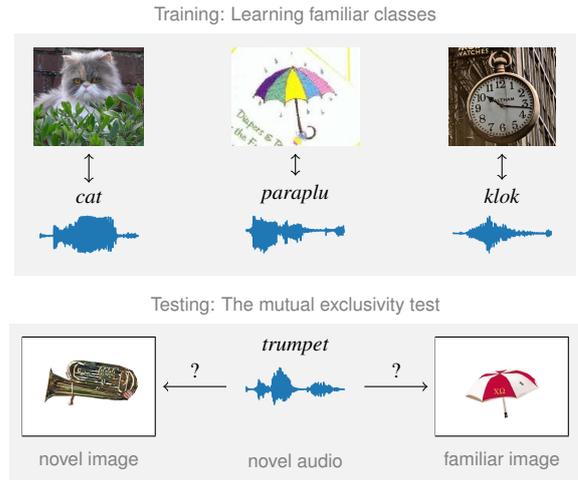

In this study we investigate the bilingual ME bias from a computational perspective using multimodal machine learning models.
While the ME bias in bilingual learners has not been studied computationally, there are studies on the monolingual bias. %
Most of these use text--image models associating %
written words with visual objects; findings are mixed in reproducing 
the ME bias~\cite{gandhi_mutual_2020,gulordava_deep_2020,vong_cross-situational_2022}.
Very recently, ME has been studied in visually grounded speech (VGS) models~\cite{nortje_visually_2024-1}, which better approximate children's reliance on spoken (rather than written) language by operating on images and spoken words.
Moreover, \cite{nortje_visually_2024-1} reliably reproduces monolingual children's ME bias.

Taking Nortje \etal~\cite{nortje_visually_2024-1} as a starting point, we compare the ME bias in monolingual and bilingual VGS models.
We first improve the model, leading to better and more consistent results in the monolingual case.
We then train various combinations of monolingual and bilingual VGS models using English, French and Dutch speech--image data, as illustrated in Fig.~\ref{fig:overview}.
We find that in most cases the bilingual models exhibit a smaller ME bias, but the results are not consistent (as in the human studies).

We then present analyses to try and understand the organisation of the resulting VGS embedding spaces in mono- and bilingual models.
In both cases, we find a modality gap in the joint audio--image space. Qualitatively, we also show for the first time that novel concepts are placed in-between familiar concepts in the embedding space.
The main difference between monolingual and bilingual models is that the familiar images are packed tighter in the bilingual case.
Our analyses lay the foundation for future work at larger scales on more language pairs.

\section{Data and method}

To investigate the ME bias computationally, we use a VGS model that takes audio and images as input. We %
train the model to associate spoken utterances of object names to their visual correspondences (Fig.~\ref{fig:overview}-top).
The concepts seen during training are then familiar to the learner.
To test the model's ME bias after learning, we prompt it with a spoken query from an unseen novel class and ask it to select one of the two images, one showing a familiar and the other a novel object (Fig.~\ref{fig:overview}-bottom).
If the model tends to associate the novel spoken word with the novel object, it has an ME bias.

The setup is identical in the mono- and bilingual cases, except that in the bilingual setting, the training and test words come from two languages.
For the bilingual English--Dutch example in Fig.~\ref{fig:overview}, the novel \textit{trumpet} (bottom) is not seen in either language during training, but the familiar \textit{umbrella}\,/\,\textit{paraplu} is seen in both English and Dutch during training.

Below we first describe how we construct the bilingual speech--image datasets used for training the VGS models.
We then describe how the model is structured, trained, and evaluated.

\subsection{Multilingual datasets}
\label{sec:datasets}

We need a dataset containing images paired with spoken words in multiple languages.
We therefore extend the English speech--image data from \cite{nortje_visually_2024-1} with Dutch and French speech.

The English data from \cite{nortje_visually_2024-1} contains 13 familiar word classes and 20 novel word classes.
These are concrete nouns like the ones in Fig.~\ref{fig:overview}: \word{cat}, \word{umbrella}, \word{clock}.
Speech segments for these words are sourced from %
the Flickr Audio Captions Corpus~\cite{harwath_deep_2015},
Buckeye~\cite{pitt_mark_buckeye_2005}, and
LibriSpeech~\cite{panayotov_librispeech_2015}.
The corresponding images come from 
MS\,COCO~\cite{lin_microsoft_2014},
Caltech-101~\cite{fei-fei_one-shot_2006}, and
ImageNet~\cite{krizhevsky_imagenet_2017}.
For training, full images are used,
while the test images contain the objects isolated using a white background mask based on the object segmentations provided with the datasets.
Training therefore happens in a cluttered natural environment, while testing is done through unambiguous evaluations; this is similar to children learning a language in a naturalistic setting and then being tested in a laboratory.

We obtain Dutch and French spoken words for the 33 classes present in the English data.
Dutch words are extracted from the Corpus Gesproken Nederlands~\cite{oostdijk_spoken_2000} and the Dutch subsets of multilingual LibriSpeech~\cite{pratap_mls_2020} and Common Voice~\cite{ardila_common_2020}. 
For French, we use %
the subsets from multilingual LibriSpeech and Common Voice.
We isolate the target spoken words using forced alignment \cite{mcauliffe_montreal_2017,pratap_mms_2024}.
Since for one novel word, \word{nautilus},
we have few samples in Dutch and French, we discard it, leaving us with 19 novel words.
The class distribution follows the source data,
with roughly 67k images and 4k audio samples for the most common concept (\word{dog})
and 47 images and 107 audio samples for the least common concept (\word{scissors})
across the three languages.

\subsection{Visually grounded speech (VGS) model architecture}
\label{sec:model}

Our VGS model simulates word learning in a cognitively motivated manner, similar to a child learning to map object names to their visual referents. Given 
an audio $\ab$ and image $\ib$, the model produces a score $\phi(\ab, \ib)$ of how well the two match.

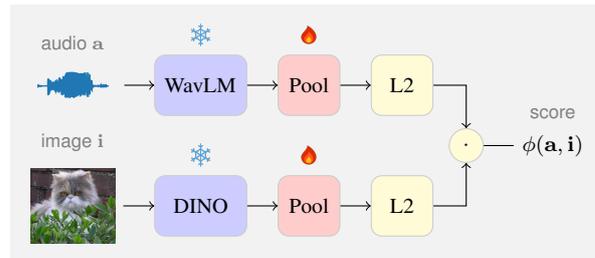
\begin{figure}[!t]
    \centering
    \input{fig-architecture}
    \caption{%
        The architecture of our VGS model.
        The only parameters that are updated are the transformer pooling layers.
    }
    \label{fig:architecture}
\end{figure}

As illustrated in %
Fig.~\ref{fig:architecture},
the architecture of the VGS model is
a two-tower encoder network coupled with a contrastive loss.
We use
WavLM~\cite{chen_wavlm_2022} to extract features from the audio $\ab$, and
DINO~\cite{caron_emerging_2021} to extract features from the image $\ib$.
Both %
have been pretrained on other unlabelled datasets using
self-supervised learning, and are kept frozen throughout VGS model training.
This can be seen as a proxy for how children during word learning can use visual and auditory perceptual abilities previously acquired from exposure in the respective modalities~\cite{clark_how_2004,zhuang_unsupervised_2021}.
The feature extractors return sequences of representations;
to obtain a single embedding vector for each modality, we use a pooling layer. %
The resulting embeddings are then L2 normalised.
A dot product gives
the similarity score $\phi(\ab, \ib)$ between the two inputs.

The pooling layer %
consists of a single transformer block with a learnable CLS
token as the query vector.
This layer also incorporates two down-projection layers:
at the input, from the feature dimension $D$
to the transformer width $W$, and
at the output, from the transformer width $W$ to the embedding dimension~$E$.
Based on validation experiments, we use transformer blocks with $W = E = 256$, four heads ($W / 64$), and an inner MLP dimension of 1024 ($W \times 4$). %
The two transformer blocks (one for audio and one for image) are the only learnable layers in the model, amounting
to approximately 2.5M parameters.

Compared to the VGS model %
from~\cite{nortje_visually_2024-1},
our architecture is both more modern (updates the AlexNet image encoder to ResNet and the CPC audio encoder to WavLM)
and simpler (pools both modalities in the same way instead of pooling only the audio modality and then max-pooling the scores).
These changes substantially improve the model's ability to discriminate between familiar classes (as we show below in Sec.~\ref{sec:experiments}), allowing us to look for the ME bias in strong learners.
Finally, since we rely on frozen encoders, training is also more efficient.

\subsection{Model training}
At each training step, the model receives an audio--image pair $(\ab^+, \ib^+)$ corresponding to the same word class and a set of negative audio samples $\left\{\ab^-\right\}$ and image samples $\left\{\ib^-\right\}$ from other classes.
Both the positive and negative samples come from familiar classes.
The negatives are sampled independently for the two modalities, so they are not necessarily matched.
We define the probability that the model matches the spoken word $\ab^+$ to the correct image $\ib^+$ as follows:
\begin{equation}
    p(\ib^+ | \ab^+) = \frac{%
        \exp\left\{\phi(\ab^+, \ib^+)/\tau\right\}}{%
        \sum_{\ib} \exp\left\{\phi(\ab^+, \ib)/\tau\right\}%
    },
\end{equation}
where $\ib$ in the denominator ranges across the positive $\ib^+$ and negative $\ib^-$ samples, and $\tau$ is a learnable temperature parameter~\cite{radford_learning_2021}.
We define the reverse conditional probability $p(\ab^+ | \ib^+)$ analogously and optimise the parameters (the pooling layer and the temperature $\tau$) to maximise the log of these two probabilities averaged across the samples in a batch.

The model is trained for 24 epochs using a learning rate %
with a linear warm-up for the first four epochs up to a learning rate of $2 \times 10^{-4}$,
followed by cosine annealing to $10^{-6}$.
In an epoch we go through all audio samples in the dataset, and for each audio sample we randomly pick a positive image and 11 negative images.
We monitor the performance on a validation split and select %
the model with the lowest loss on this split.
For the encoders, we use the \texttt{base-plus} WavLM variant~\cite{chen_wavlm_2022} (pretrained on English data; $D = 768$)
and the ResNet-50 DINO variant~\cite{caron_emerging_2021} ($D = 2048$).
The temperature $\mathbf{\tau}$ is capped at 100.

We train monolingual (English, Dutch, and French) and bilingual (English--Dutch, English--French, and Dutch--French) VGS models on the 13 familiar classes from our speech--image datasets. %
Since the number of epochs is fixed and in each epoch we go through all the audio samples,
the bilingual models will go through more updates than the monolingual models.
\new{Results were similar when an equal number of training steps was used.}

\subsection{Model evaluation}

To test a model, we present it with an audio query and two images:
a positive one, which matches the class of the audio query, and
a negative one, belonging to a different class from the audio query.
Using this protocol, we consider two types of tests. %

\textit{Familiar test}.
First, we measure discrimination ability across familiar classes to ensure the models are properly trained.
In this test,
the audio query belongs to a familiar class, and both images are also from familiar classes (one matching and one~not).

\textit{ME test}. Second, we quantify the ME bias. In this test, the audio query and the positive image come from a novel class, and the negative image from a familiar class  \new{(as in Fig.~\ref{fig:overview}-bottom)}.

In both tests, we sample 50 episodes for each class from the test set, 
\new{with}
no overlap between 
\new{train,}
validation, and test samples.
To prevent dataset biases, the two images in each pair are drawn from the same source %
dataset~(Sec.~\ref{sec:datasets}). 
Each experiment is repeated five times with a different
\new{seed affecting weight initialisation and data sampling.}
For the monolingual models, the test audio query matches the training language. For bilingual models, the audio query can come from either training~language.

\section{Results and analyses}
\label{sec:experiments}

We want to see whether both monolingual and bilingual models show an ME bias; whether the strength of the bias differs between the two; and what the similarities and differences are in how the embedding spaces are organised.

\subsection{Experimental results}
\label{sec:results}

The results in Table~\ref{tab:main-results} are ordered according to the language of the test query.
Familiar performance is close to perfect in all settings, for all languages, and even in the bilingual case.
The models have therefore properly learned the familiar words and can robustly recognise the corresponding objects despite the masked-out background at test time.
This validates our computational setup and is crucial for a meaningful ME test.
Note that overall, our results on the familiar test substantially improve on those reported in prior work~\cite{nortje_visually_2024-1}, likely due to the use of self-supervised features in our model (see Sec.~\ref{sec:model}).

Having established that the models are well-trained, we now consider the ME test.
All the monolingual models yield an accuracy of 67--68\%, so they all show an ME bias (%
50\% would indicate no bias).
These results show for the first time consistent ME biases for monolingual VGS models trained on languages other than English.
Our scores \new{on English} are higher than the ME results of roughly 60\% in~\cite{nortje_visually_2024-1}, suggesting that stronger learners (like the model in this study) show a stronger ME bias.

\begin{table}[!t]
\caption{
    Performance on the familiar and ME tests
    for monolingual and bilingual VGS models.
    We report mean accuracy (\%) and standard error computed across five training seeds.
}
\label{tab:main-results}
\renewcommand{\arraystretch}{1.1}
\begin{tabularx}{\linewidth}{@{}
    L
    c
    r
    r
    @{}}
    \toprule
    \multicolumn{1}{@{}l}{Training languages} & 
    \multicolumn{1}{c}{Test language} & \multicolumn{1}{c}{Familiar} & \multicolumn{1}{c}{ME} \\
    \midrule
    Monolingual: EN   &              & 99.4±0.1 & 66.2±1.1 \\
    Bilingual: EN, FR & English (EN) & 99.6±0.1 & 65.7±1.3 \\
    Bilingual: EN, NL &              & 99.6±0.1 & 63.5±1.5 \\
    \midrule
    Monolingual: FR   &              & 98.5±0.4 & 67.6±1.4 \\
    Bilingual: FR, EN &  French (FR) & 98.9±0.1 & 66.8±1.4 \\
    Bilingual: FR, NL &              & 99.0±0.1 & 69.4±0.9 \\
    \midrule
    Monolingual: NL    &             & 98.5±0.3 & 67.3±1.3 \\
    Bilingual: NL, EN  & Dutch (NL)  & 98.7±0.3 & 63.5±2.1 \\
    Bilingual: NL, FR  &             & 98.6±0.3 & 65.7±1.2 \\
    \bottomrule
\end{tabularx}
\end{table}

We now turn to our main question: a comparison of the ME results between monolingual and bilingual models.
When tested on English, we see that the ME score of 66.2\% for an English-only model drops to 65.7\% when French is added, or to 63.5\% when Dutch is added (Table~\ref{tab:main-results} top).
ME scores similarly drop when English is added on the French test (middle), and when English or French is added on the Dutch test (bottom).
This trend matches the findings from experiments on children, indicating that bilingual children make less use of the ME bias than monolingual children~\cite{byers2009, byers2017, houston2010}.
But our result does not hold in every single case: the Dutch--French model tested on French gives an ME score of 69.4\%, which is higher than the 67.6\% from the French-only model.

The results in Table~\ref{tab:main-results} are for models with roughly 2.5M parameters, but we also repeated the experiments with smaller ($W = 128$, 0.8M parameters) and larger models ($W = 512$, 
8.2M parameters).
\new{
We observe similar trends: in most but not all cases the bilingual models exhibit a smaller ME bias compared to the monolingual ones.
We carried out several statistical comparisons between bilingual and their corresponding monolingual models. Out of 18 tests (3 languages $\times$ 2 language pairs $\times$ 3 model sizes), 9 show that the bias is significantly stronger in the bilingual models, and 6 show a difference in the expected direction but not significantly so. Human studies have similar discrepancies: %
it isn't always the case that bilingual children show less ME, with
age \cite{davidson2005, kalashnikova_effects_2015}, vocabulary \cite{byers2013}, and the type of test \cite{davidson1997, kalashnikova_effects_2015} all seeming to play a role.}

We have shown that visually grounded bilingual models tend to have a lower ME bias than monolingual ones.
But why does this happen?
Below we first look at why the ME bias is seen in general, before considering the mono- vs bilingual question.

\subsection{Understanding the embedding space}

In an analysis of English-only models, Nortje \etal~\cite{nortje_visually_2024-1} showed quantitatively that novel audio samples are generally closer to novel images (regardless of class) than they are to any familiar class.
This is why we see the ME bias.
But how is the representation space organised qualitatively?
The structure of the model of~\cite{nortje_visually_2024-1} did not allow for a sensible visualisation of the embedding spaces,%
\footnote{The model in \cite{nortje_visually_2024-1} did not extract an image embedding, but rather patch embeddings that were aggregated non-linearly into the final score.}
but our changes (Sec.~\ref{sec:model}) enable visualisation.
We therefore contribute the following new analyses.

\begin{figure}[!t]
    \centering
    \includegraphics[width=\linewidth]{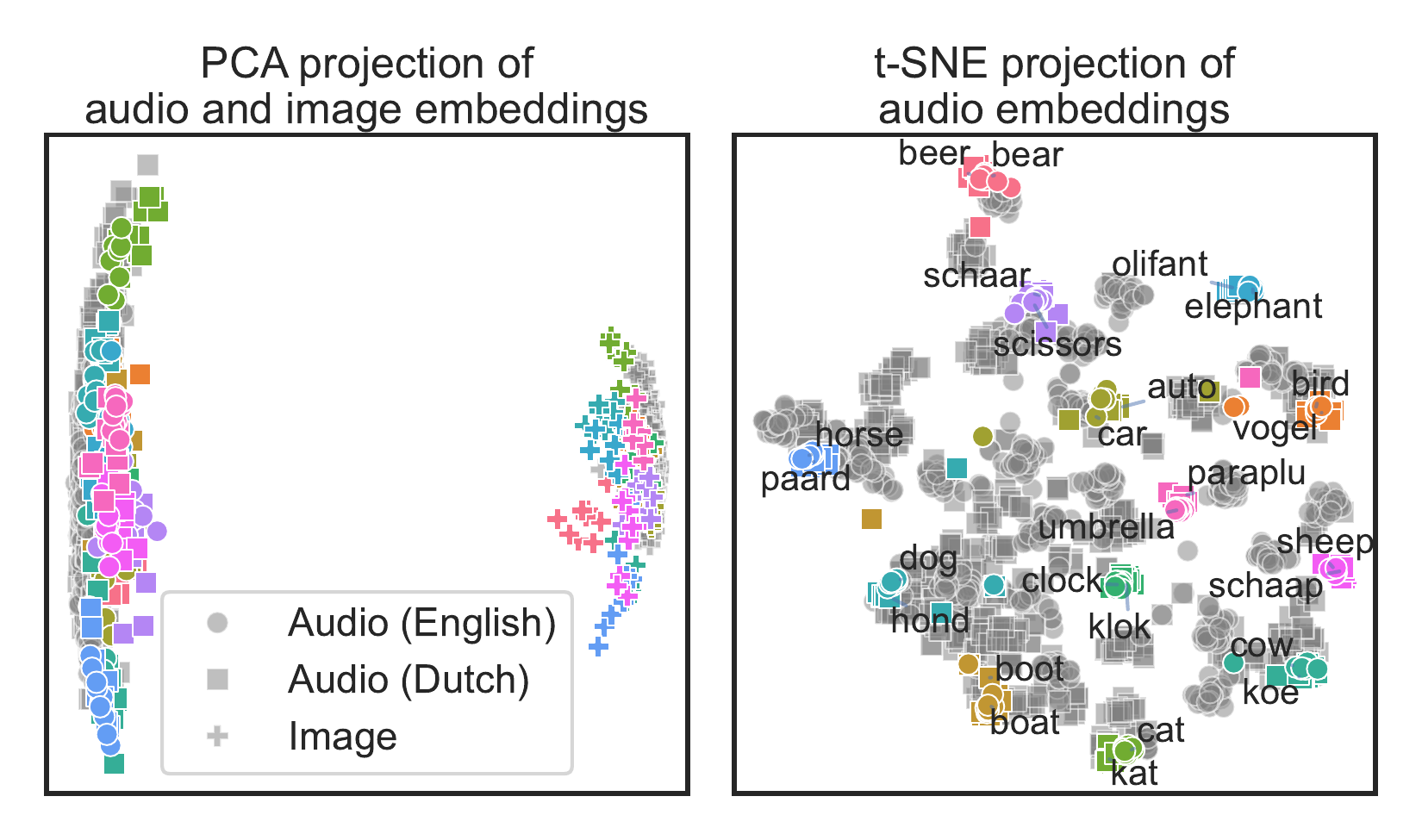}
    \caption{%
        Projections of embeddings from the bilingual English--Dutch model.
        Familiar classes are coloured; novel classes are grey.
        Left: Audio and image embeddings live in different cones of the unit sphere.
        Right: Bilingual audio embeddings are aligned for the familiar classes, while novel classes live in-between.
    }
    \label{fig:embeddings}
\end{figure}

Fig.~\ref{fig:embeddings}-left uses a %
PCA projection to visualise the 256-dimensional embeddings for both the audio and image samples.
The linear PCA projection allows us to see the global structure of the data.
Since the embeddings are L2 normalised, they live on the unit sphere.
The audio and image embeddings are positioned on different sides.
This is known as the modality gap and has been observed in other bimodal models %
trained with a contrastive loss~\cite{liang_mind_2022}.
Despite the gap, the familiar audio and image classes are well-aligned, e.g., the green points are at the top and the blue points at the bottom for both modalities.
This ought to be the case for the strong familiar
results in Table~\ref{tab:main-results}.

\begin{figure}[!t]
    \centering
    \includegraphics[width=\linewidth]{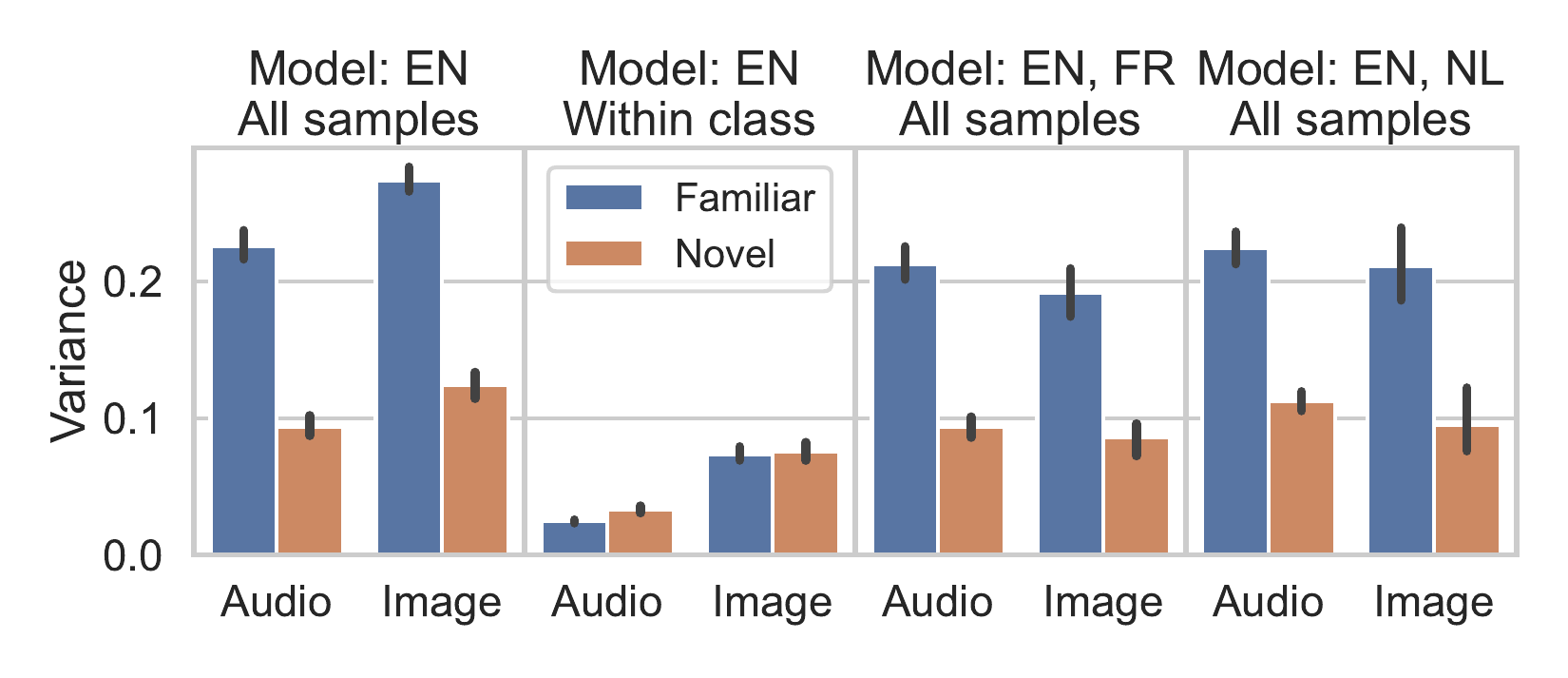}
    \caption{%
        Variance of familiar and novel samples for three models computed across all samples or samples within each class.
        Familiar samples occupy a larger space than novel ones (first plot),
        but they are tighter within class than novel samples (second plot).
        In the bilingual models (third and fourth plots), the image space gets tighter than in the monolingual case.
        \new{Error bars are 95\% confidence intervals computed by bootstrapping.}
    }
    \label{fig:variance}
\end{figure}

The PCA picture shows global structure but
is not complete: from quantitative measurement, we know that on average novel audio (grey circles and squares) are closer to novel images (grey crosses) than they are to familiar images (coloured crosses), but this is not captured in Fig.~\ref{fig:embeddings}-left.
We therefore analyse individual modalities using a non-linear visualisation.
Fig.~\ref{fig:embeddings}-right
uses t-SNE to visualise the audio embeddings.
The model separates out familiar classes throughout the representation space and places novel classes in an in-between region.
We verify this interpretation quantitatively by computing the variance of familiar and novel samples, both when all the samples are combined or when considering samples per class.%
\footnote{The variance is the trace of the covariance matrix,
or, equivalently, the mean of the distances between samples and their centroid.}
Fig.~\ref{fig:variance} shows results for an English model and two bilingual models.
We see that, indeed, the overall variance of the novel data is much smaller than that of the familiar data.
At the same time, the per-class variance, shown in the second plot in Fig.~\ref{fig:variance}, is {somewhat} smaller for the familiar than for the novel data.
Taken together, these variance values support
the hypothesis stated in~\cite{nortje_visually_2024-1} that during training the model spreads familiar classes around the space, with each class in a tight bundle of its own, while novel classes are placed between familiar ones, in a region where novel samples from different classes overlap.
Fig.~\ref{fig:variance} shows that this is the case for both mono- and bilingual models.

\subsection{Monolingual vs bilingual models}

The analyses above give new insight into why we observe the ME bias in both mono- and bilingual models.
But why is the bias slightly weaker in bilingual models? %
Given that ME is the result of comparisons between two modalities in a high-dimensional space (with a modality gap), it is inherently difficult to 
explain
the small differences in results.
\new{But} we can use the variance analysis of Fig.~\ref{fig:variance} to get some insights.
A consistent change when adding either French or Dutch to the English model is that the spread across samples in the visual modality becomes tighter, in particular for familiar samples (compare the blue image bars in the first plot vs the third and fourth plots).
While the variance of all the novel images is also slightly smaller (brown bars),
we speculate that
the familiar space shrinks more (relatively speaking), resulting  
in more novel items being confused with familiar ones (i.e., a lower ME bias) in the bilingual case.

This is not a comprehensive analysis: the ME bias exists because of comparisons across modalities, and not within a modality (which is what we do in the analysis here).
Differences are also small, which complicates analyses.
Repeating our analysis at a larger scale %
will therefore be necessary in future~work.

\subsection{Can a bilingual model implicitly translate?}

An interesting secondary research question is how the bilingual models structure their acoustic spaces given supervision through the visual modality.
In Fig.~\ref{fig:embeddings}-right, we see that the audio embeddings of the familiar words in the two languages overlap.
This happens regardless of whether the two words sound similar (\word{clock}--\word{klok}) or not (\word{horse}--\word{paard}).
We quantify this translation performance by measuring the accuracy of a simple nearest mean centroid classifier:
for each audio sample in one language, we find its closest audio centroid in the other language. %
With this approach, we achieve translation accuracies over 97\% for all language pairs.
The accuracies for the novel words are, as expected, poor: less than 30\% in all cases. This is still better than random ($\text{5.2\%} = \text{1} / \text{19}$ novel words), because some of the words are the same in all three languages (e.g., \word{bus}, \word{piano}).

\section{Conclusion}

In this study we investigated whether the ME bias---a heuristic employed by children in language learning---is also seen in visually grounded speech models trained on bilingual speech--image data.
We found that bilingual models consistently exhibit an ME bias
and that the strength of the bias tends to be weaker than for monolingual models, with some exceptions.
These findings are consistent with those observed in children, where the ME bias has been reported to be generally lower in bilingual children, but this pattern is somewhat inconsistent \cite{byers2009, byers2017, houston2010, davidson2005, kalashnikova_effects_2015, byers2013, kalashnikova2018, davidson1997}.
\new{While the results of our computational study cannot explain why certain patterns are observed in children, our model nonetheless can be used to generate predictions that can then be tested in experiments with children.
Furthermore, in our computational model,}
we can carefully control the training data and analyse the model's internal representations.
We relied on these advantages in this paper, but plan to use it even more in future work, increasing the number of language pairs and the vocabulary size to gain even more insights into the nature of the ME bias.

\section{Acknowledgements}

\new{This work was in part supported by the EU Horizon project AI4TRUST (No. 101070190) and by CNCS-UEFISCDI (PN-IV-P7-7.1-PTE-2024-0600).}

\bibliography{ref}

\end{document}

%% file: fig-overview.tex
\def\h{1.3cm}
\def\ha{0.7cm}

\begin{tikzpicture}[
        font=\footnotesize,
        label1/.style={label distance=-1.5mm},
    ]
    \newcommand{\mylabel}[1]{{\scriptsize\color{gray}\textsf{#1}}}
    \matrix [row sep=0.1cm, column sep=0.8cm] {
        \node (tr-img-1)  {\includegraphics[height=\h , trim={1.5cm 0cm 1.5cm 0cm}, clip]{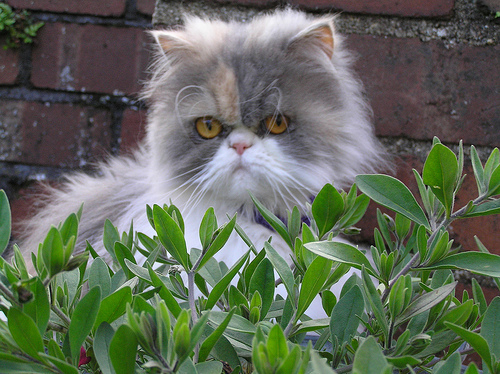}}; &
        \node (tr-img-2)  {\includegraphics[height=\h, trim={0cm 0cm 0cm 0cm}, clip]{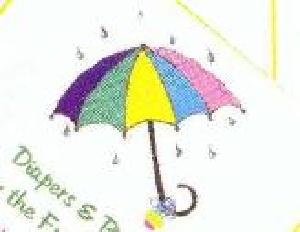}}; &
        \node (tr-img-3)  {\includegraphics[height=\h, trim={0cm 3cm 0cm 4cm}, clip]{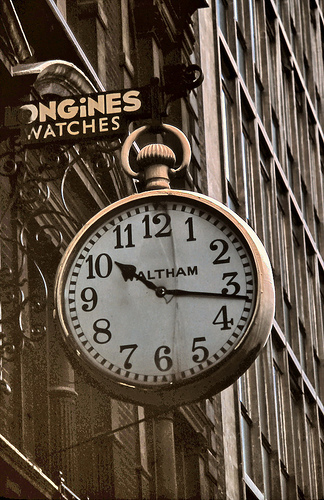}}; \\[0.2cm]
        \node[label={[label1, name=tr-label-1]\word{cat}}] (tr-audio-1)  {\includegraphics[height=\ha]{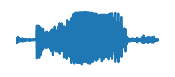}}; &
        \node[label={[label1, name=tr-label-2]\word{paraplu}}] (tr-audio-2)  {\includegraphics[height=\ha]{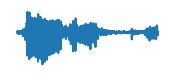}}; &
        \node[label={[label1, name=tr-label-3]\word{klok}}] (tr-audio-3)  {\includegraphics[height=\ha]{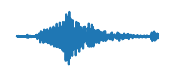}}; \\[0.7cm]
        \node (pos)       {\frame{\includegraphics[height=\h, trim={0cm 0cm 0cm 0cm}, clip]{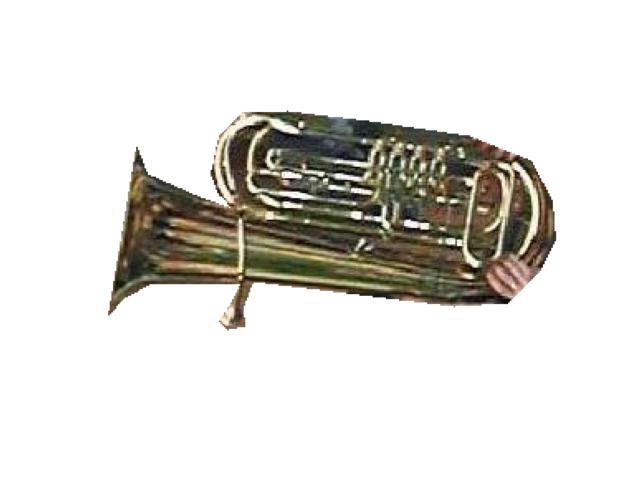}}}; &
        \node[align=center, label={[label1]\word{trumpet}}] (query)     {\includegraphics[height=\ha]{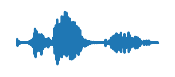}}; &
        \node (neg)       {\frame{\includegraphics[height=\h, trim={0cm 0cm 0cm 0cm}, clip]{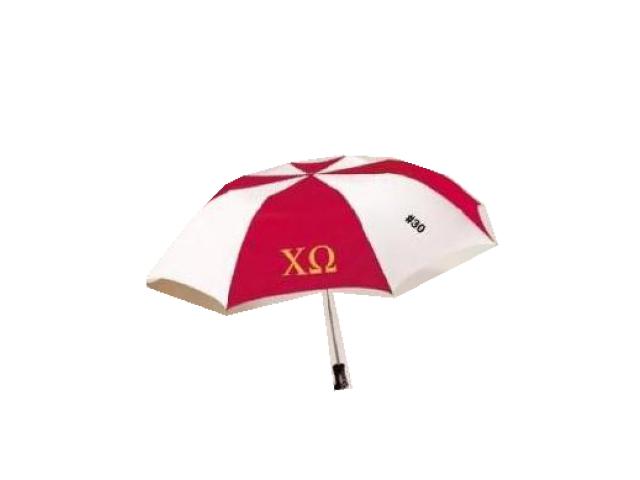}}}; \\[-0.1cm]
        \node (label-pos)   {\mylabel{novel image}}; &
        \node (label-query) {\mylabel{novel audio}}; &
        \node (label-neg)   {\mylabel{familiar image}}; \\
    };

    \draw[<->] (tr-img-1) -- (tr-label-1);
    \draw[<->] (tr-img-2) -- (tr-label-2);
    \draw[<->] (tr-img-3) -- (tr-label-3);

    \draw[->] (query) -- node [above] {?} (pos);
    \draw[->] (query) -- node [above] {?} (neg);
    
    \begin{pgfonlayer}{background}
        \node [
            label=\mylabel{Training: Learning familiar classes},
            fill=gray!10,
            fit=(tr-img-1) (tr-img-2) (tr-img-3) (tr-audio-1) (tr-audio-2) (tr-audio-3),
            inner sep=2pt,
        ] {};
    \end{pgfonlayer}
    
    \begin{pgfonlayer}{background}
        \node [
            label=\mylabel{Testing: The mutual exclusivity test},
            fill=gray!10,
            fit=(pos) (query) (neg) (label-pos) (label-query) (label-neg),
            inner sep=2pt,
        ] {};
    \end{pgfonlayer}
\end{tikzpicture}

%% file: fig-architecture.tex
\begin{tikzpicture}[
        font=\footnotesize,
        block/.style={draw=black!20, minimum height=0.8cm, rounded corners, align=center},
        frozen/.style={fill=blue!20},
        trainable/.style={fill=red!20},
        noparams/.style={fill=yellow!20},
        tw1/.style={text width=1cm},
        tw2/.style={text width=0.6cm},
        proj/.style={trapezium, trapezium angle=60, minimum width=1.5mm, fill=red!20, draw=black!20},
        stop/.style={draw, shape=circle, fill=black, minimum size=0.1cm, inner sep=0pt},
        port/.style={rectangle, minimum width=0.5cm, inner sep=0pt},
    ]
    \newcommand{\mylabel}[1]{{\scriptsize\color{gray}\textsf{#1}}}
    \newcommand{\tlabel}[1]{{\tiny\color{gray}\textsf{#1}}}
    \begin{scope}[local bounding box=overview]
        \matrix [row sep={0.8cm, between origins}, column sep=0.4cm] {
            \node[label=\mylabel{audio $\ab$}] (audio) {\includegraphics[height=0.5cm]{imgs/fig-overview/audio-cat_6272-70191-0029.png}}; &
            \node[%
                block,
                tw1,
                frozen,
                label={\includegraphics[height=0.3cm]{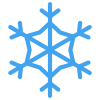}}
            ] (audio-enc) {WavLM}; &
            \node[
                block,
                tw2,
                trainable,
                label={\includegraphics[height=0.3cm]{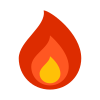}}
            ] (audio-pool) {Pool}; &
            \node[block, tw2, noparams] (audio-norm) {L2}; & [-0.2cm] & \\
            & & & & 
            \node[circle, noparams, draw=black!20] (dot-prod) {$\cdot$}; &
            \node[label=\mylabel{score}] (score) {$\phi(\ab, \ib)$}; \\
            \node[label=\mylabel{image $\ib$}] (image) {\includegraphics[height=1cm, trim={1.5cm 0cm 1.5cm 0cm}, clip]{imgs/fig-overview/cat_n02123394_981.jpg}}; &
            \node[
                block,
                tw1,
                frozen,
                label={\includegraphics[height=0.3cm]{imgs/fig-architecture/icons8-snowflake-100.png}}
            ] (image-enc) {DINO}; &
            \node[
                block,
                tw2,
                trainable,
                label={\includegraphics[height=0.3cm]{imgs/fig-architecture/icons8-fire-100.png}}
            ] (image-pool) {Pool}; &
            \node[block, tw2, noparams] (image-norm) {L2}; \\
        };
    
        \draw [->] (audio) -- (audio-enc);
        \draw [->] (audio-enc) -- (audio-pool);
        \draw [->] (audio-pool) -- (audio-norm);
        \draw [->] (audio-norm) -| (dot-prod);
        
        \draw [->] (image) -- (image-enc);
        \draw [->] (image-enc) -- (image-pool);
        \draw [->] (image-pool) -- (image-norm);
        \draw [->] (image-norm) -| (dot-prod);
    
        \draw (dot-prod) -- (score);
    \end{scope}

    \begin{pgfonlayer}{background}
        \node [
            fill=gray!10,
            fit=(overview),
            inner sep=1pt,
        ] {};
    \end{pgfonlayer}

\end{tikzpicture}